\documentclass[12pt,journal,draftcls,a4paper,twoside,onecolumn]{IEEEtran}

\usepackage{setspace}
\usepackage{setspace}
\usepackage{dsfont}
\usepackage{amsmath,amsthm}
\usepackage{amssymb}
\usepackage{amsfonts}
\usepackage{graphicx}
\usepackage[all,poly]{xy}
\usepackage{multirow}
\usepackage{epstopdf}
\usepackage{color}
\usepackage[noadjust]{cite}
\usepackage{subfigure}
\usepackage{mathrsfs}
\usepackage{algorithmic}
\usepackage[linesnumbered,lined,ruled]{algorithm2e}
\usepackage{cancel}
\usepackage{epsfig}
\usepackage{stfloats}
\usepackage{placeins}
\usepackage{float}
\usepackage{url}

\newfloat{bottomtwocolumnequation}{b}{fff}
\floatname{bottomtwocolumnequation}{Equation}

\newtheorem{lemma}{Lemma}

\newcommand{\CirEig}{\bfD}
\newcommand{\bs}{\boldsymbol}

\newcommand{\NoiCovMat}{\bs{\Lambda}}
\newcommand{\tcpp}[1]{\tcp{\emph{\small{#1}}}}
\newcommand{\undS}{\underline{\bfS}}

\newcommand{\vecD}{\bar{\bfD}}

\newcommand{\Emp}[1]{{\textcolor[rgb]{1.00,0.00,0.00}{{#1}}}}

\def\bfc{{\mathbf{c}}}

\def\bfu{{\mathbf{u}}}

\def\bfA{{\mathbf{A}}}
\def\bfB{{\mathbf{B}}}
\def\bfC{{\mathbf{C}}}
\def\bfD{{\mathbf{D}}}

\def\bfF{{\mathbf{F}}}

\def\bfH{{\mathbf{H}}}

\def\bfJ{{\mathbf{J}}}

\def\bfL{{\mathbf{L}}}
\def\bfM{{\mathbf{M}}}
\def\bfN{{\mathbf{N}}}

\def\bfP{{\mathbf{P}}}
\def\bfQ{{\mathbf{Q}}}

\def\bfS{{\mathbf{S}}}

\def\bfU{{\mathbf{U}}}

\def\bfX{{\mathbf{X}}}
\def\bfY{{\mathbf{Y}}}

\def\calO{{\mathcal{O}}}

\def\wtm{\widetilde{m}}

\newcommand{\MATima}{\bfX}

\newcommand{\nbbandima}{m_{\lambda}}

\newcommand{\argmin}{\mathrm{arg}\min}

\newcommand{\Vzeros}[1]{\boldsymbol{0}_{#1}}
\newcommand{\Id}[1]{\textbf{I}_{#1}}

\newcounter{algo}
\renewcommand{\thealgo}{\arabic{algo}} 

\graphicspath{{figures/}}

\begin{document}
\title{R-FUSE: Robust Fast Fusion of Multi-Band Images Based on Solving a Sylvester Equation}

\author{\IEEEauthorblockN{Qi Wei, \IEEEmembership{Member,~IEEE},
Nicolas Dobigeon, \IEEEmembership{Senior Member,~IEEE}, \\
Jean-Yves
Tourneret, \IEEEmembership{Senior Member,~IEEE}, Jos\'e Bioucas-Dias, \IEEEmembership{Member,~IEEE} 
and Simon Godsill, \IEEEmembership{Senior Member,~IEEE}\\}
\thanks{Part of this work has been supported by the ERA-NET MED MapInvPlnt Project no. ANR-15-NMED-0002-02, by the Hypanema ANR Project no. ANR-12-BS03-003
and by ANR-11-LABX-0040-CIMI within the program
ANR-11-IDEX-0002-02 within the thematic trimester on image processing. }
\thanks{ Qi Wei and Simon Godsill are with Department of Engineering,
University of Cambridge, CB2 1PZ, Cambridge, UK (e-mail: \{qi.wei, sjg\}@eng.cam.ac.uk).}
\thanks{
Nicolas Dobigeon and Jean-Yves Tourneret are with University of Toulouse, IRIT/INP-ENSEEIHT, 2 rue
Camichel, BP 7122, 31071 Toulouse cedex 7, France (e-mail: \{nicolas.dobigeon, jean-yves.tourneret\}@enseeiht.fr).}
\thanks{
Jos\'e Bioucas-Dias is with Instituto de Telecomunica\c{c}\~oes and
Instituto Superior T\'ecnico, Universidade de Lisboa, Portugal (e-mail: \{bioucas\}@lx.it.pt).
}}

\maketitle
\vspace{-1cm}

\begin{abstract}
This paper proposes a robust fast multi-band image fusion method to merge
a high-spatial low-spectral resolution image and a low-spatial high-spectral
resolution image. Following the method recently developed in \cite{Wei2015FastFusion},
the generalized Sylvester matrix equation associated with the multi-band image fusion problem is solved in a 
more robust and efficient way by exploiting the Woodbury formula, avoiding
any permutation operation in the frequency domain as well as the 
blurring kernel invertibility assumption required in \cite{Wei2015FastFusion}.
Thanks to this improvement, the proposed algorithm requires fewer computational operations 
and is also more robust with respect to the blurring kernel compared with the one in \cite{Wei2015FastFusion}. 
The proposed new algorithm is tested with different priors considered in \cite{Wei2015FastFusion}.
Our conclusion is that the proposed fusion algorithm
is more robust than the one in \cite{Wei2015FastFusion} with a reduced computational cost.
\end{abstract}

 \vspace{-0.2cm}
\section{Introduction}
\label{sec:intro}
\subsection{Background}
\label{subsec:background}

The multi-band image fusion problem, i.e., fusing hyperspectral
(HS) and multispectral (MS)/panchromatic (PAN) images 
has recently been receiving particular attention in remote sensing \cite{Loncan2015}. 
High spectral resolution multi-band imaging generally suffers from the limited spatial
resolution of the data acquisition devices, mainly due to an unavoidable tradeoff between spatial and spectral sensitivities \cite{Chang2007}. 
For example, HS images benefit from excellent spectroscopic properties with hundreds of bands but are
limited by their relatively low spatial resolution compared to MS and PAN images (that are acquired in much fewer bands). As a consequence, reconstructing a high-spatial and high-spectral multi-band image from two degraded and complementary observed images is a challenging but crucial issue that has been addressed in various scenarios {\cite{Aanaes2008,Beyerera2011Bayesian,Gong2012change,James2014medical}}. In particular, fusing a high-spatial low-spectral resolution image and a low-spatial high-spectral image is an archetypal instance of multi-band image reconstruction, such as pansharpening (MS+PAN) \cite{Aiazzi2012} or hyperspectral pansharpening (HS+PAN) \cite{Loncan2015}.
Generally, the linear degradations applied to the observed images with respect
to (w.r.t.) the target high-spatial and high-spectral image reduce to
spatial and spectral transformations. Thus, the multi-band image fusion
problem can be interpreted as restoring a three dimensional data-cube from two
degraded data-cubes. A detailed formulation of this problem is presented in the next section.
 \vspace{-0.2cm}
\subsection{Problem Statement}
\label{subsec:prob_stat}

This work directly follows the formulation of \cite{Wei2015FastFusion} and uses the well-admitted linear degradation model
\begin{equation}
\begin{array}{ll}
\label{eq:obs_general}
\bfY_{\mathrm{L}} =  {\bfL} \MATima + \bfN_{\mathrm{L}}\\
\bfY_{\mathrm{R}} =  \MATima {\bf BS} + \bfN_{\mathrm{R}}
\end{array}
\end{equation}
where
\begin{itemize}
\item $\bfX \in\mathbb{R}^{m_{\lambda} \times n}$ is the full resolution target image,
\item ${\bfY}_{\mathrm{L}}\in\mathbb{R}^{n_{\lambda} \times n}$ and ${\bfY}_{\mathrm{R}} \in \mathbb{R}^{\nbbandima \times m}$
are the observed spectrally degraded and spatially degraded images,
\item ${\bfL} \in\mathbb{R}^{n_{\lambda} \times \nbbandima}$ is the spectral response of the sensor,
\item $\bfB \in \mathbb{R}^{n\times n}$ is a cyclic convolution operator on the bands, \item $\bfS \in \mathbb{R}^{n\times m}$ is a $d=d_r \times d_c$ uniform downsampling operator, which has $m=n/d$ ones on the block diagonal and zeros elsewhere, such that ${\bfS}^H\bfS=\Id{m}$,
\item $\bfN_{\mathrm{L}}\in\mathbb{R}^{n_{\lambda} \times n}$ and $\bfN_{\mathrm{R}}\in \mathbb{R}^{\nbbandima \times m}$ are additive noise terms that are assumed to be distributed according to the following matrix normal
distributions \cite{Gupta1999}
\begin{equation*}
    \begin{array}{ll}
      \bfN_{\mathrm{L}}  \sim \mathcal{MN}_{\nbbandima,m}(\Vzeros{\nbbandima,m},\NoiCovMat_{\mathrm{L}} ,\Id{m} ) \\
      \bfN_{\mathrm{R}}  \sim \mathcal{MN}_{n_{\lambda},n}(\Vzeros{n_{\lambda},n}, \NoiCovMat_{\mathrm{R}}, \Id{n}).
    \end{array}
\end{equation*}
\end{itemize}

Computing the ML or the Bayesian estimators (associated with any prior) is a challenging task, mainly due to the large size of $\bfX$ and to 
the presence of the downsampling operator $\bfS$, which prevents any direct use of the Fourier transform (FT) to diagonalize the 
the joint spatial degradation operator $\bf{BS}$. To overcome this difficulty, several computational strategies have been designed to approximate the estimators. Based 
on a Gaussian prior, a Markov chain Monte Carlo (MCMC) algorithm was implemented in \cite{Wei2015JSTSP} to generate a collection
of samples asymptotically distributed according to the posterior distribution of $\bfX$. The Bayesian estimators of $\bfX$ can then be approximated using these samples. Despite this formal appeal, MCMC-based methods have the major drawback of being computationally expensive, which prevents their effective use when processing images of large size. Relying on exactly the same prior model, the strategy developed in \cite{Wei2015whispers} exploits an alternating direction method of multipliers (ADMM) embedded in a block coordinate descent method (BCD) to compute the maximum a posterior (MAP) estimator of $\bfX$. This optimization strategy allows the numerical complexity to be greatly decreased when compared to its MCMC counterpart. Based on a prior built from a sparse representation, the fusion problem was solved in \cite{Simoes2015,Wei2015TGRS} with the split augmented Lagrangian shrinkage algorithm (SALSA) \cite{Afonso2011}, which is an instance of ADMM. In \cite{Wei2015FastFusion}, contrary to the algorithms described above, a much more efficient method was proposed to solve explicitly an underlying Sylvester equation (SE) derived from \eqref{eq:obs_general}, {leading to an algorithm referred to as Fast fUsion based on Sylvester Equation (FUSE)}. This {algorithm} can be implemented \emph{per se} to compute the ML estimator in a computationally efficient manner. The proposed {FUSE algorithm} has also the great advantage of being easily generalizable within a Bayesian framework when considering various priors. The MAP estimators associated with a Gaussian prior similar to \cite{Wei2015JSTSP,Wei2015whispers} can be directly computed thanks to the proposed strategy.
When handling more complex priors such as \cite{Simoes2015,Wei2015TGRS}, the {FUSE} solution can be conveniently embedded within a conventional ADMM or a BCD algorithm. Although the FUSE algorithm has significantly accelerated the fusion of multi-band images, it requires the non-trivial assumption that the blurring matrix $\bfB$ is invertible, which is not always guaranteed in practice. 

In this work, we propose a more robust version of FUSE algorithm, which is termed as R-FUSE. In this R-FUSE, the FT of the target image, instead of its blurring version as in \cite{Wei2015FastFusion}, is computed explicitly by exploiting the Woodbury formula. A direct consequence of this modification is getting rid of the invertibility assumption
for the blurring matrix $\bfB$. A side product of this modification is that the permutations conducted in the frequency domain (characterized by the matrix $\bfP$ 
in \cite{Wei2015FastFusion}) are no longer required in the R-FUSE algorithm.
 \vspace{-0.3cm}
\section{Problem Formulation}
\label{sec:ObsModel}
Since adjacent HS bands are known to be highly correlated, the columns of $\bfX$
usually reside in a subspace whose dimension $\wtm_{\lambda}$ is much smaller than the number
of bands $\nbbandima$ \cite{Chang1998,Bioucas2008}, i.e., $\bf X = HU$
where $\bfH$ is {a full column rank matrix and}
$\bfU \in \mathbb{R}^{\wtm_{\lambda} \times n}$ is the projection of $\bfX$ onto the subspace
spanned by the columns of $\bfH \in \mathbb{R}^{\nbbandima  \times \wtm_{\lambda}}$.

According to the maximum likelihood or least squares (LS) principles,
the fusion problem associated with the linear model \eqref{eq:obs_general} 
can be formulated as 
\begin{equation}
\label{eq:neglog_likeli}
\argmin\limits_{\bfU} L(\bfU)
\end{equation}
where
\begin{equation*}
L(\bfU) = \| \NoiCovMat_{\mathrm{R}}^{-\frac{1}{2}}\left(\bfY_{\mathrm{R}}- \bf{HUBS}\right)\|_F^2 +
\| \NoiCovMat_{\mathrm{L}}^{-\frac{1}{2}}\left(\bfY_{\mathrm{L}} -{{\bfL}\bf HU}\right)\|_F^2
\end{equation*}
and $\|\cdot\|_F$ represents the Frobenius norm.
 \vspace{-0.2cm}
\section{Robust Fast Fusion Scheme}
\label{sec:FastFus}
\subsection{Sylvester equation}
\label{subsec:opt_ima}
Minimizing \eqref{eq:neglog_likeli} w.r.t. $\bfU$ is equivalent
to force the derivative of $L(\bfU)$ to be zero, i.e.,$\nabla L(\bfU)=0$, 
leading to the following matrix equation
\begin{equation}
\label{eq:sylvester}
\begin{array}{ll}
\bfH^H \NoiCovMat_{\mathrm{R}}^{-1} {\bf{HUBS}} \left(\bf BS \right)^H + \left({\left({\bfL}\bfH\right)}^H \NoiCovMat_{\textrm{L}}^{-1} {{\bfL}\bfH}\right) \bfU \\
= \bfH^H \NoiCovMat_{\mathrm{R}}^{-1} {\bfY}_\mathrm{R} \left(\bf BS \right)^H + {\left({\bfL}\bfH\right)}^H \NoiCovMat_{\textrm{L}}^{-1} {\bfY}_\mathrm{L}.\\
\end{array}
\end{equation}
As mentioned in Section \ref{subsec:prob_stat}, the difficulty for solving \eqref{eq:sylvester} results from
the high dimensionality of $\bfU$ and the presence of the downsampling matrix $\bfS$. The work in \cite{Wei2015FastFusion} 
showed that Eq. \eqref{eq:sylvester} can be solved analytically with two assumptions
\begin{itemize}
\item The blurring matrix $\bfB$ is a block circulant matrix with circulant blocks.
\item The decimation matrix $\bfS$ corresponds to downsampling the original image and its conjugate transpose $\bfS^H$ interpolates the decimated image with zeros.
\end{itemize}

As a consequence, the matrix $\bfB$ can be decomposed as $\bfB = \bfF \CirEig \bfF^H$ with
$\bfB^H = \bfF \CirEig^{\ast} \bfF^H$, where $\bfF \in \mathbb{R}^{n\times n}$
is the discrete Fourier transform (DFT) matrix ($\bfF\bfF^H=\bfF^H\bfF=\Id{n}$), $\bfD \in \mathbb{R}^{n\times n}$
is a diagonal matrix and $\ast$ represents the conjugate operator. 
Another non-trivial assumption used in \cite{Wei2015FastFusion} is that 
the matrix $\bfD$ (or equivalently $\bfB$) is invertible, which is not necessary 
in this work as shown in Section \ref{subsec:r-fuse}.
The decimation matrix satisfies the property $\bfS^H\bfS=\Id{m}$ and the matrix $\undS \triangleq
\bfS\bfS^H \in \mathbb{R}^{n\times n}$ is symmetric and idempotent, i.e., $\undS=\undS^H$ and $\undS \undS^H=\undS^2=\undS$.
For a practical implementation, multiplying an image by $\undS$ can be achieved by doing entry-wise multiplication with
an $n\times n$ mask matrix with ones in the sampled position and zeros elsewhere.

After multiplying \eqref{eq:sylvester} on both sides by $\left(\bfH^H \NoiCovMat_{\mathrm{R}}^{-1} \bfH\right)^{-1}$,
we obtain
\begin{equation}
\label{eq:zeroforce}
\bfC_1 \bfU +  \bfU \bfC_2  = \bfC
\end{equation}
where\\
$\bfC_1=\left(\bfH^H \NoiCovMat_{\mathrm{R}}^{-1} \bfH\right)^{-1}\left({\left({\bfL}\bfH\right)}^H \NoiCovMat_{\textrm{L}}^{-1} {{\bfL}\bfH}\right)$\\
$\bfC_2={\bfB \undS \bfB}^H$\\
\vspace{-0.3cm}
\begin{equation}
\bfC=\left(\bfH^H \NoiCovMat_{\mathrm{R}}^{-1} \bfH\right)^{-1} \left(\bfH^H \NoiCovMat_{\mathrm{R}}^{-1} {\bfY}_\mathrm{R} \left(\bf BS \right)^H + {\left({\bfL}\bfH\right)}^H\NoiCovMat_{\textrm{L}}^{-1} {\bfY}_\mathrm{L} \right).
\label{eq:Compute_C}
\end{equation}
Eq. \eqref{eq:zeroforce} is a Sylvester matrix equation that admits a unique solution if and only if an arbitrary sum of the eigenvalues
of $\bfC_1$ and $\bfC_2$ is not equal to zero \cite{Bartels1972}.
 \vspace{-0.5cm}
\subsection{Proposed closed-form solution}
\label{subsec:r-fuse}
Using the eigen-decomposition $\bfC_1=\bfQ{\bf{\Lambda}}\bfQ^{-1}$ and multiplying both sides
of \eqref{eq:zeroforce} by $\bfQ^{-1}$ leads to
\begin{equation}
{\bf{\Lambda}} \bfQ^{-1} \bfU + \bfQ^{-1}\bfU \bfC_2  = \bfQ^{-1} \bfC.
\label{eq:sylv_1}
\end{equation}
Right multiplying \eqref{eq:sylv_1} by the DFT matrix $\bfF$ on both sides and using the definitions of matrices $\bfC_2$ and $\bf B$ yields
\begin{equation}
{\bf{\Lambda}} \bfQ^{-1} {\bf{UF}}+ \bfQ^{-1}{\bf{UF}}\left( \bfD\bfF^H \undS{\bfF} \bfD^{\ast}  \right) = \bfQ^{-1} \bfC \bf F.
\label{eq:sylv_2}
\end{equation}
Note that ${\bf{UF}} \in \mathbb{R}^{\wtm_{\lambda} \times n}$ is the
FT of the target image, which is a complex matrix. Eq. \eqref{eq:sylv_2} can be regarded
as an SE w.r.t. $\bfQ^{-1} {\bf{UF}}$, which has a simpler form compared to \eqref{eq:zeroforce}
as ${\bf{\Lambda}}$ is a diagonal matrix. Instead of using any block permutation matrix as in \cite{Wei2015FastFusion},
we propose to solve the SE \eqref{eq:sylv_2} row-by-row (i.e., band-by-band).
Recall the following lemma originally proposed in \cite{Wei2015FastFusion}.
\begin{lemma}[Wei \em{et al.}, \cite{Wei2015FastFusion}]
The following equality holds
\begin{equation}
{\bfF}^H \undS {\bfF} = \frac{1}{d}\bfJ_{d} \otimes \Id{m} 
\end{equation}
where ${\bfF}$ and $\undS$ are defined as in Section \ref{subsec:opt_ima}, $\bfJ_d$ is the $d \times d$ matrix of
ones and $\Id{m}$ is the $m \times m$ identity matrix.
\label{lemm:WEI}
\end{lemma}
By simply decomposing the matrix $\bfJ_{d}$ as $\bfJ_{d}=\bs{1}_d \bs{1}_d^T$, where $\bs{1}_d \in \mathbb{R}^d$ is a vector of ones
and using the mixed-product property of Kronecker product, i.e., $\left(\bfA_1 \bfA_2\right) \otimes\left(\bfA_3 \bfA_4\right)=\left(\bfA_1 \otimes\bfA_3\right) \left(\bfA_2 \otimes\bfA_4\right)$ (if $\bfA_1$, $\bfA_2$, $\bfA_3$ and $\bfA_4$ are matrices of proper sizes), we can easily get the following result
\begin{equation}
{\bfF}^H \undS {\bfF} = \frac{1}{d}(\bs{1}_d \otimes \Id{m})(\bs{1}_d^T \otimes \Id{m})
\label{eq:Lemma_2}
\end{equation}
Substituting \eqref{eq:Lemma_2} into \eqref{eq:sylv_2} leads to
\begin{equation}
{\bf{\Lambda}} \bar{\bfU} + \bar{\bfU}\bfM  = \bar{\bfC}
\label{eq:Sylv_3}
\end{equation}
where $\bar{\bfU}={\bfQ^{-1}\bf{UF}}, \bfM =\frac{1}{d}\vecD\vecD^H, 
\vecD=\bfD\left(\bs{1}_d \otimes \Id{m}\right), \bar{\bfC} ={\bfQ^{-1}\bfC}{\bfF}$.
Eq. \eqref{eq:Sylv_3} is an SE w.r.t. $\bar{\bfU}$ whose solution is
significantly easier than the one of \eqref{eq:sylv_1}, due to the 
simple structure of the matrix $\bfM$. To ease the notation, the diagonal 
matrices $\bf{\Lambda}$ and $\bfD$ are rewritten as 
${\bf{\Lambda}}=\textrm{diag}\left(\lambda_1,\cdots,\lambda_{\wtm_{\lambda}}\right)$ 
and $\bfD=\textrm{diag}\left(\bfD_1,\cdots,\bfD_d\right)$,
where $\textrm{diag}\left(\cdot_1,\cdots,\cdot_k\right)$ represents a 
(block) diagonal matrix whose (block) diagonal elements are $\cdot_1,\cdots,\cdot_k$
and $\lambda_i \geq 0$, $\forall i$. Thus, we have $\vecD^H\vecD=\sum\limits_{t=1}^d\bfD_t^H\bfD_t=\sum\limits_{t=1}^d\bfD_t^2$.

In the following, we will show that \eqref{eq:Sylv_3} can be solved row-by-row explicitly.
First, we rewrite $\bar{\bfU}$ and $\bar{\bfC}$ as $\bar{\bfU}=\left[\bar{\bfu}_1^T, \cdots ,
\bar{\bfu}_{\wtm_{\lambda}}^T\right]^T$ and $\bar{\bfC}=\left[\bar{\bfc}_1^T, \cdots, \bar{\bfc}_{\wtm_{\lambda}}^T\right]^T$, 
where $\bar{\bfu}_i \in \mathbb{R}^{1\times n}$ and $\bar{\bfc}_i \in \mathbb{R}^{1\times n}$
are row vectors. Using these notations, \eqref{eq:Sylv_3} can be decomposed as 
\begin{equation*}
\lambda_i \bar{\bfu}_i + \bar{\bfu}_i \bfM = \bar{\bfc}_i
\end{equation*}
for $i=1,\cdots,\wtm_{\lambda}$.
Direct computation leads to 
\begin{equation}
\bar{\bfu}_i= \bar{\bfc}_i \left( \bfM+\lambda_i \Id{n}\right)^{-1}.
\label{eq:syv_band}
\end{equation}
Following the Woodbury formula \cite{Woodbury1950} and using $\vecD^H\vecD=\sum\limits_{t=1}^d\bfD_t^2$,
the inversion in \eqref{eq:syv_band} can be easily computed as 
$\left(\bfM+\lambda_i \Id{n}\right)^{-1}=\lambda_i^{-1}\Id{n}-\lambda_i^{-1}\bar{\bfD}\left(\lambda_i d \Id{m}+\sum\limits_{t=1}^d\bfD_t^2\right)^{-1}\bar{\bfD}^H$.
As $\lambda_i d \Id{m}+\sum\limits_{i=1}^d\bfD_i^2$ is a real diagonal matrix, its inversion is easy to be computed with a complexity of order $\calO\left(m\right)$.
Using this simple inversion, the solution $\bar{\bfU}$ of the SE \eqref{eq:Sylv_3} can be computed row-by-row (band-by-band) as 
\begin{equation}
\bar{\bfu}_i= \lambda_i^{-1}\bar{\bfc}_i - \lambda_i^{-1}\bar{\bfc}_i \bar{\bfD}\left(\lambda_i d \Id{m}+\sum\limits_{t=1}^d\bfD_t^2\right)^{-1}\bar{\bfD}^H
\label{eq:sol_band}
\end{equation}
for $i=1,\cdots,\wtm_{\lambda}$.
The final estimator of $\bfX$ is obtained as 
\begin{equation*}
\hat{\bfX}={\bf HQ} \bar{\bfU} {\bfF}^H.
\end{equation*}

\IncMargin{1em}
\begin{algorithm}[h!]
\label{Algo:Sylvester_Algo}
\KwIn{$\bfY_{\mathrm{L}}$, $\bfY_{\mathrm{R}}$, $\NoiCovMat_{\mathrm{L}}$,
 $\NoiCovMat_{\mathrm{R}}$, $\bfL$, $\bfB$, $\bfS$, $\bfH$, $d$}
$\bf{D} \leftarrow  \textrm{EigDec} \left(\bfB\right)$; \tcpp{FFT transformation}
\Emp{$\vecD \leftarrow \bfD\left(\bs{1}_d \otimes \Id{m}\right)$}\;
$\bfC_1 \leftarrow \bfC_1\left(\bfH,\bfL, \NoiCovMat_{\mathrm{L}},\NoiCovMat_{\mathrm{R}}\right)$;  \tcpp{Compute cf. {\eqref{eq:Compute_C}}}
$\left({\bfQ,\bf{\Lambda}}\right) \leftarrow \textrm{EigDec}\left(\bfC_1\right)$;  \tcpp{cf. $\bfC_1=\bfQ{\bf{\Lambda}}\bfQ^{-1}$}
$\bfC \leftarrow \bfC\left(\bfH,\bfL, \NoiCovMat_{\mathrm{L}},\NoiCovMat_{\mathrm{R}},\bfY_{\mathrm{L}},\bfY_{\mathrm{R}},\bfB,\bfS\right)$;  \tcpp{cf. {\eqref{eq:Compute_C}}}
$\bar{\bfC} \leftarrow {\bfQ^{-1}\bfC}{\bfF}$\;
\tcpp{Compute $\bar{\bfU}$ \Emp{band by band} ($\wtm_{\lambda}$ bands)}
 \Emp{\For{$l=1$ \KwTo $\wtm_{\lambda}$}{
$\bar{\bfu}_i \leftarrow \bar{\bfu}_i \left(\lambda_i,d,\bar{\bfc}_i,\bar{\bfD},\bfD\right) $; \tcpp{cf. \eqref{eq:sol_band}} 
}}
\Emp{Set $\hat{\bfX}= {\bf HQ}\bar{\bfU}{\bfF}^H$}\;
\KwOut{$\hat{\bfX}$}
\caption{Robust Fast fUsion based on solving a Sylvester Equation (R-FUSE)}
\DecMargin{1em}
\end{algorithm}
\vspace{-0.7cm}
\subsection{Difference with \cite{Wei2015FastFusion}}
It is interesting to mention some important differences between the proposed R-FUSE strategy and the one of \cite{Wei2015FastFusion}:
\begin{itemize}
\item The matrix $\bfB$ (or $\bfD$) is not required to be invertible.
\item Each band can be restored as a whole instead of block-by-block ($d$ blocks). 
\end{itemize}

{Algorithm} \ref{Algo:Sylvester_Algo} summarizes the derived R-FUSE steps required to
calculate the estimated image $\hat{\bfX}$, where the different parts with \cite{Wei2015FastFusion}
have been highlighted in red.
 \vspace{-1cm}
\subsection{Complexity Analysis}
The most computationally expensive part of the proposed algorithm is the computation of the matrix $\bar{\bfC}$
(because of the FFT and iFFT operations), which has a complexity of order $\mathcal{O}(\wtm_{\lambda}n\log n)$. 
The left matrix multiplications with $\bfQ^{-1}$ (to compute $\bar{\bfC}$) and with $\left(\bfH^H \NoiCovMat_{\mathrm{R}}^{-1} \bfH\right)^{-1}$ (to compute $\bfC$)
have a complexity of order $\mathcal{O}(\wtm_{\lambda}^2 n)$. Thus, the calculation 
of $\bar{\bfC}$ has a total complexity of order $\mathcal{O}(\wtm_{\lambda}n \cdot \mathrm{max}\left\{\log n,\wtm_{\lambda}\right\})$,
which can be approximated by $\mathcal{O}(\wtm_{\lambda}n\log n)$ as $\log n \gg \wtm_{\lambda}$.

Note that the proposed R-FUSE scheme can be embedded within an ADMM or a BCD algorithm to deal with Bayesian estimators, 
as explained in \cite{Wei2015FastFusion}.
\vspace{-0.2cm}
\section{Experimental results}
\label{sec:simu}
This section applies the proposed fusion method to two 
Bayesian fusion schemes (with appropriate priors for the unknown matrix $\bfX$)
that have been investigated in \cite{Wei2015whispers} and \cite{Simoes2015}.
Note that these two methods require
to solve a minimization problem similar to \eqref{eq:neglog_likeli}.
All the algorithms have been implemented using MATLAB R2015b on a computer with
Intel(R) Core(TM) i7-4790 CPU@3.60GHz and 16GB RAM. 
\vspace{-0.5cm}
\subsection{Fusion Quality Metrics}
\label{subsec:performance}
Following \cite{Wei2015FastFusion}, we used the restored signal-to-noise ratio (RSNR), 
the averaged spectral angle mapper (SAM), the universal image quality index (UIQI), the relative dimensionless global
error in synthesis (ERGAS) and the degree of distortion (DD) as quantitative measures to evaluate the quality of the 
fused results. The larger RSNR and UIQI, or the smaller SAM,
ERGAS and DD, the better the fusion. 
\vspace{-0.3cm}
\subsection{Fusion of Multi-band images}
\label{subsec:Fus_HS_MS}
\begin{figure}[h!]
\centering
   \subfigure{
   \includegraphics[width=0.23\textwidth]{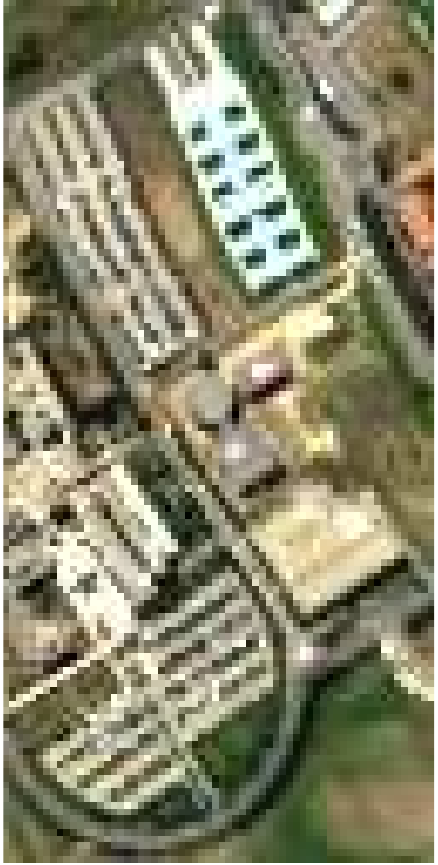}} 
   \subfigure{
   \includegraphics[width=0.23\textwidth]{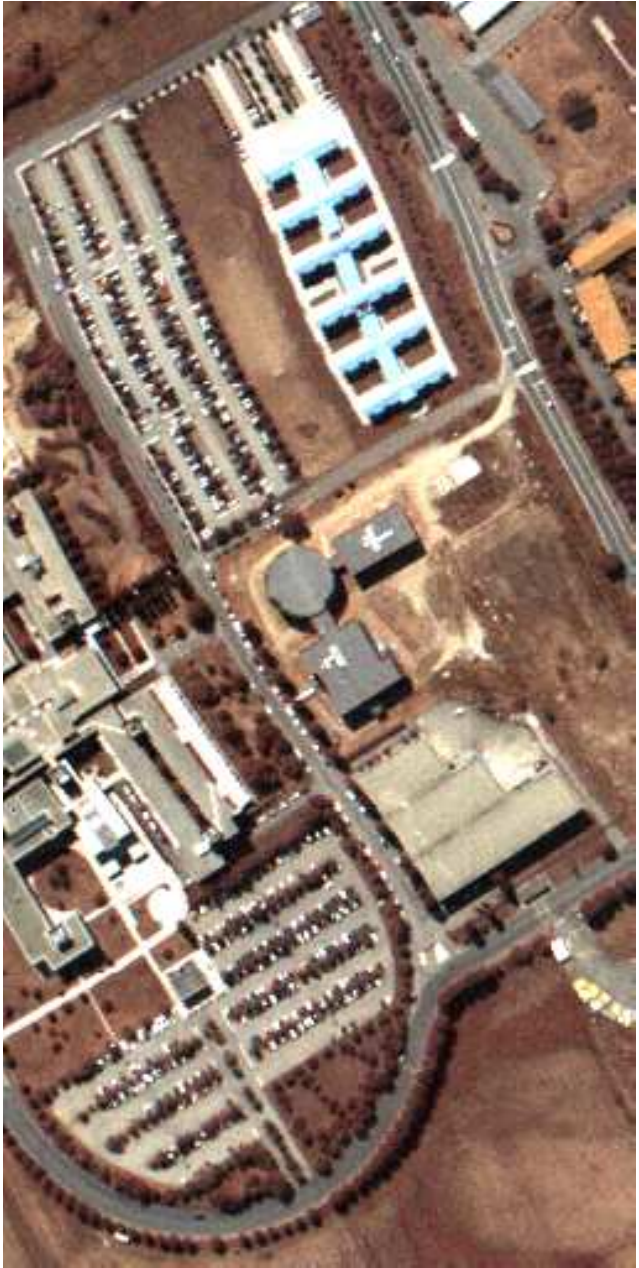}}
   \subfigure{
   \includegraphics[width=0.23\textwidth]{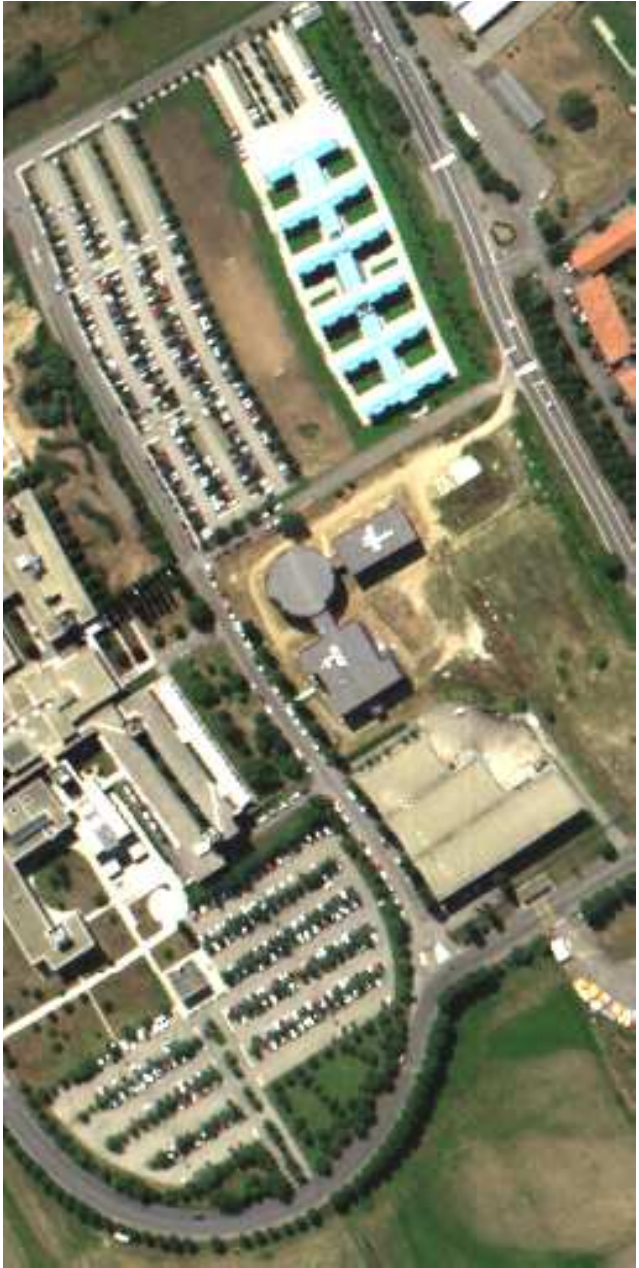}}
   \caption{Pavia dataset: HS image (left), MS image (middle) and reference image (right).}
\label{fig:observe}
\end{figure}

The reference image considered here as the high-spatial and high-spectral image is
a {$512 \times 256 \times 93$} HS image acquired over Pavia, Italy, by the reflective
optics system imaging spectrometer (ROSIS). This image was initially composed of $115$ bands
that have been reduced to $93$ bands after removing the water vapor absorption bands.
A composite color image of the scene of interest is shown in Fig. \ref{fig:observe} (right).


First, $\bfY_{\mathrm{R}}$ has been
generated by applying a $5 \times 5$ Gaussian filter (shown in the left of Figs. \ref{fig:kernel})
and by down-sampling every $d_r=d_c=4$ pixels in both vertical and horizontal directions for each band of the
reference image. Second, a $4$-band MS image $\bfY_{\mathrm{L}}$ has been obtained by
filtering $\MATima$ with the LANDSAT-like reflectance spectral responses \cite{Fleming2006}.
The HS and MS images are both contaminated by zero-mean additive Gaussian noises
with $\textrm{SNR}_{\mathrm{H}}=40$dB for HS bands and $\textrm{SNR}_{\mathrm{M}}=30$dB 
for MS bands. The observed HS and MS images are shown in Fig. \ref{fig:observe} (left and middle).

\begin{figure}[t!]
\centering
    \subfigure{
    \includegraphics[angle=0,width=0.23\textwidth,]{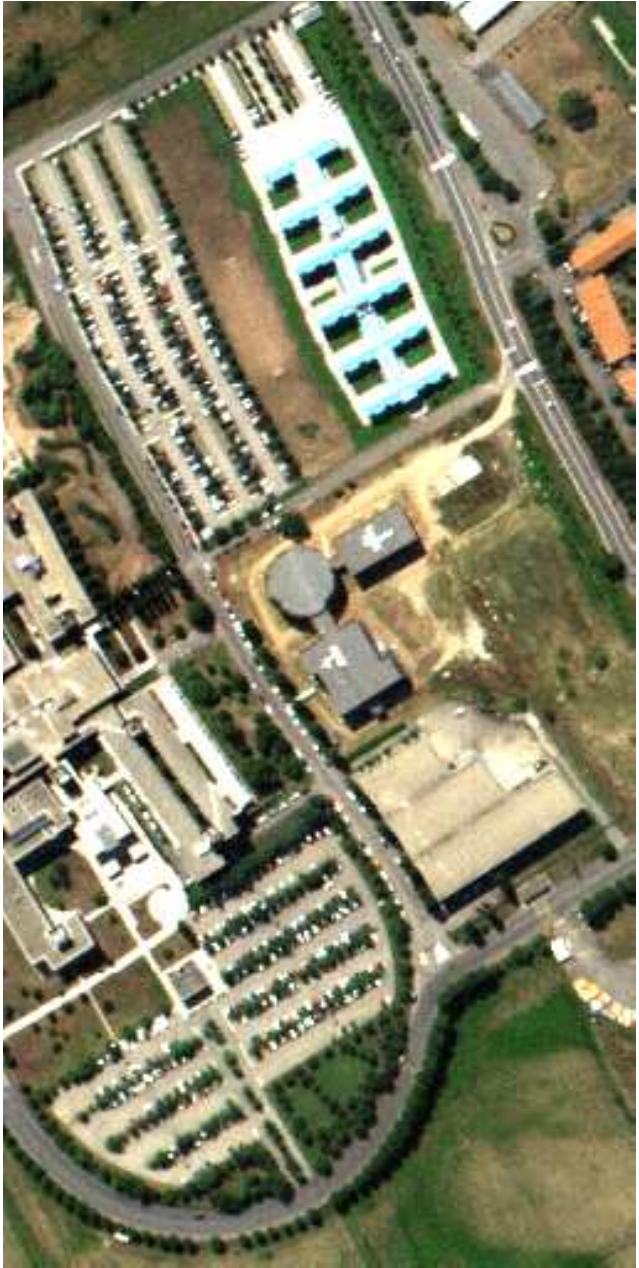}}
    \subfigure{
    \includegraphics[angle=0,width=0.23\textwidth]{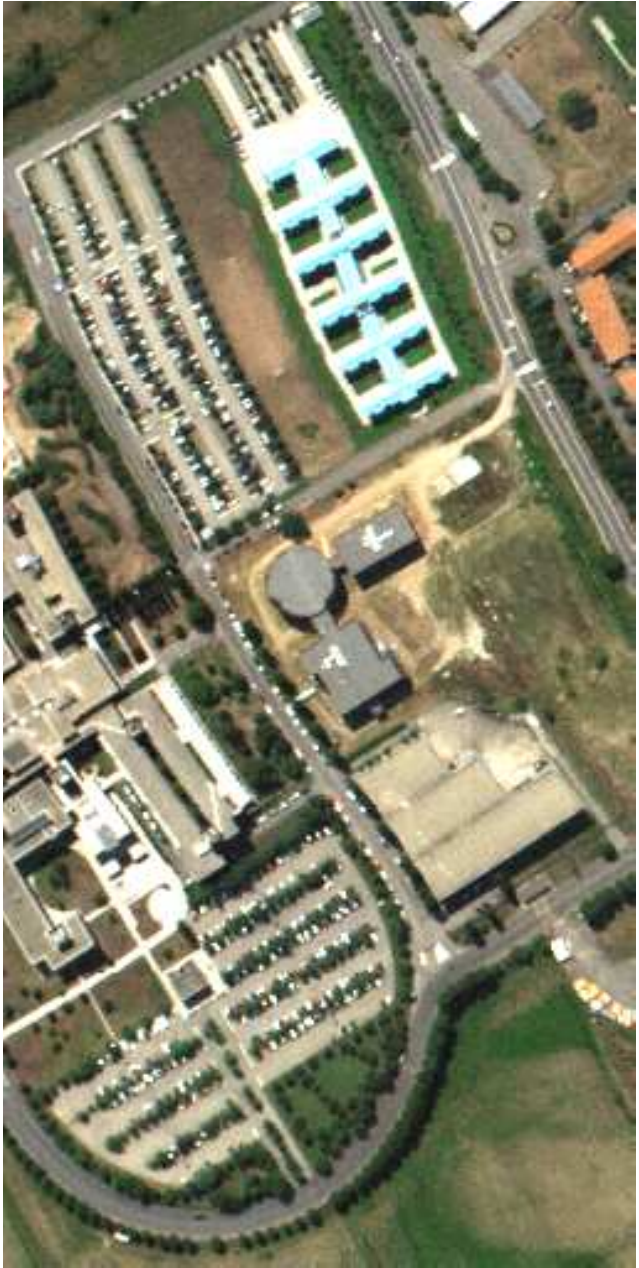}}
    \subfigure{
    \includegraphics[angle=0,width=0.23\textwidth]{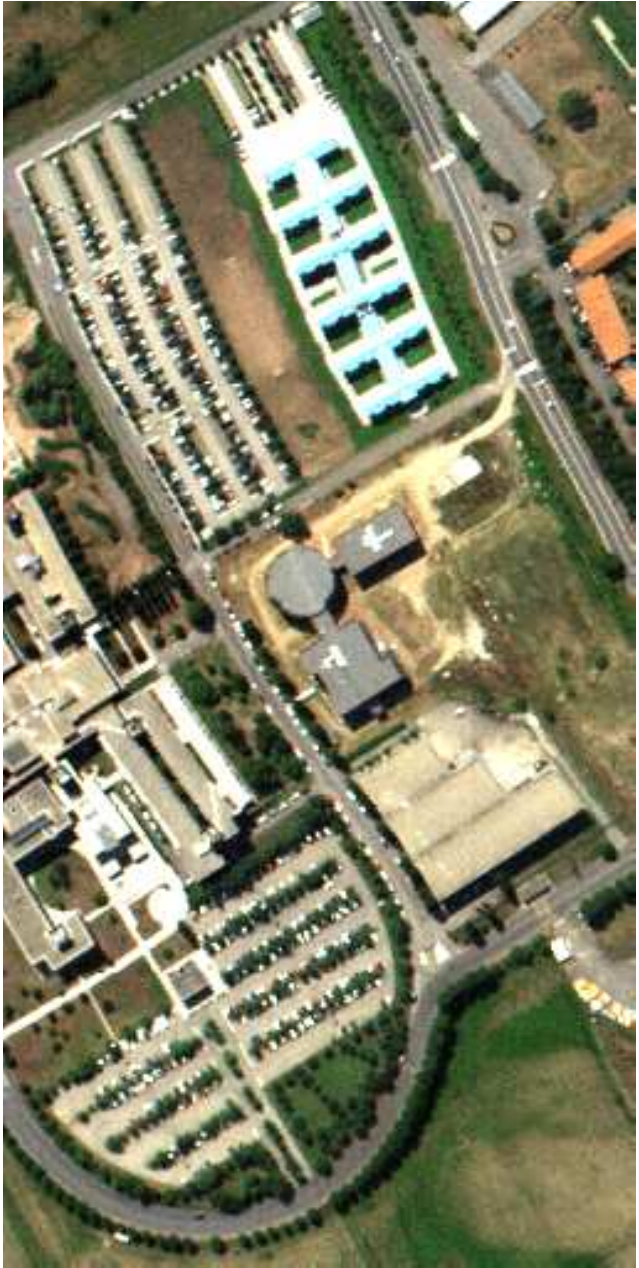}}
    \subfigure{
    \includegraphics[angle=0,width=0.23\textwidth]{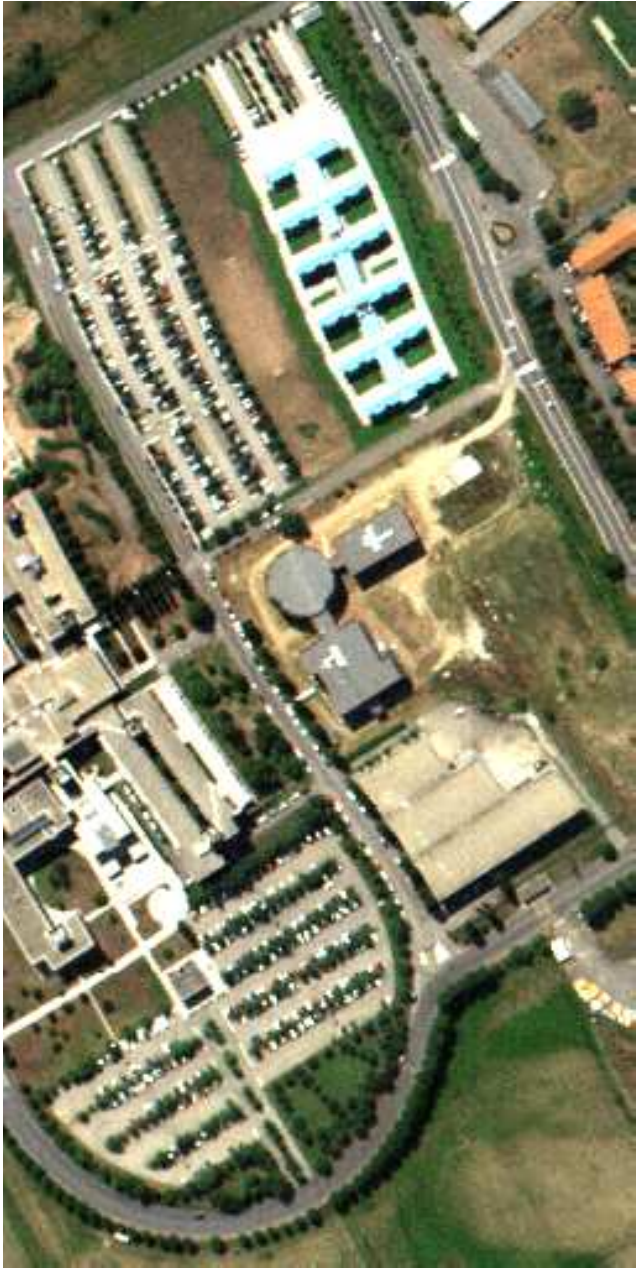}}\\
    \caption{HS+MS fusion results. 1st: FUSE using Gaussian prior, 2nd: R-FUSE using Gaussian prior,
    3rd: FUSE using TV prior and 4th: R-FUSE using TV prior.}
\label{fig:results}
\end{figure}

We consider the Bayesian fusion with Gaussian \cite{Wei2015JSTSP} and TV \cite{Simoes2015} priors that were
considered in \cite{Wei2015FastFusion}. The proposed R-FUSE and FUSE algorithms 
are compared in terms of their performance and computational time 
for the same optimization problem (corresponding to (18) in \cite{Wei2015FastFusion}). 
The estimated images obtained with the different algorithms are depicted in Fig. \ref{fig:results} and are visually very
similar. The corresponding quantitative results are reported in Table \ref{tb:quality} and confirm the same performance 
of FUSE and R-FUSE in terms of the various fusion quality measures (RSNR, UIQI, SAM, ERGAS and DD). Note that the results
associated with a TV prior are slightly better than the ones obtained with a Gaussian prior, which can be attributed to the
well-known denoising property of the TV prior.
A particularity of the R-FUSE algorithm is its reduced computational complexity
due to the avoidance of any permutation in the frequency domain when solving the Sylvester matrix
equation, as demonstrated by the computational time also reported in Table \ref{tb:quality}.

\newcommand{\figresultwidth}{0.30\textwidth}

\begin{table}[t!]
\centering
\renewcommand{\arraystretch}{1.1}
\setlength{\tabcolsep}{0.5mm}
\caption{Performance of HS+MS fusion methods: RSNR (in dB), UIQI, SAM (in degree), ERGAS, DD (in $10^{-3}$) and time (in second).}
\vspace{0.1cm}
\begin{tabular}{|c|c|cccccc|}
\hline
Prior & Methods & RSNR & UIQI & SAM  & ERGAS & DD & Time \\
\hline
\hline
\multirow{2}{*}{Gaussian} & FUSE &  29.243  & 0.9904 & 1.513 & 0.902 & 6.992 & 0.27\\
\cline{2-8}
 &R-FUSE & 29.243  & 0.9904 & 1.513 & 0.902 & 6.992 & \Emp{0.24}\\
\hline
\hline
\multirow{2}{*}{TV} &FUSE&   29.629  & 0.9914 & 1.456 & 0.853 &6.761  &133\\
\cline{2-8}
&R-FUSE &  29.629  & 0.9914 & 1.456 & 0.853 &6.761 &\Emp{115}\\
\hline
\end{tabular}
\label{tb:quality}
\end{table}

\vspace{-0.5cm}
\subsection{Robustness w.r.t. the blurring kernel}
\begin{figure}[t!]
\centering
    \subfigure{
    \includegraphics[angle=0,width=0.24\textwidth,]{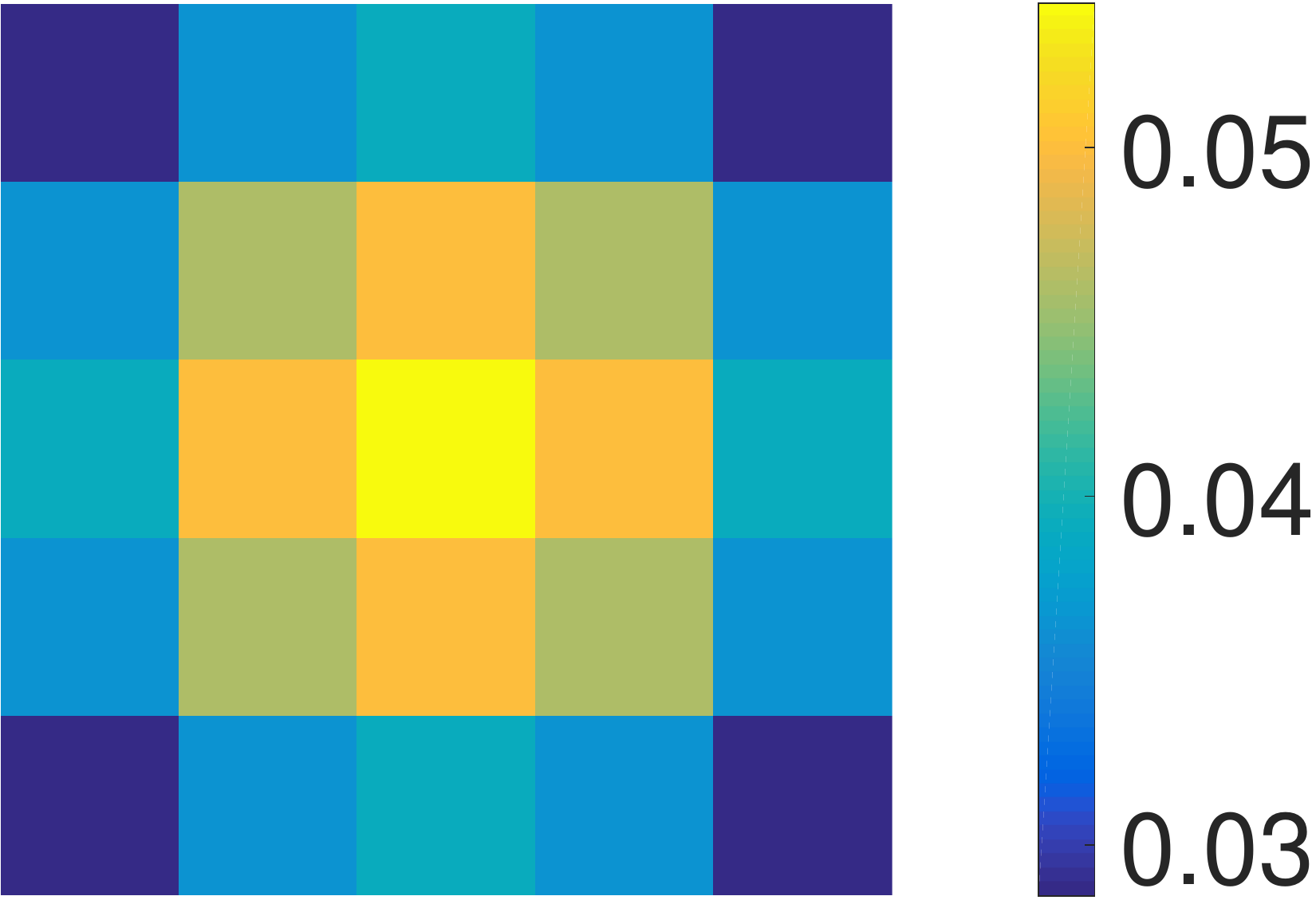}}
    \subfigure{
    \includegraphics[angle=0,width=0.24\textwidth]{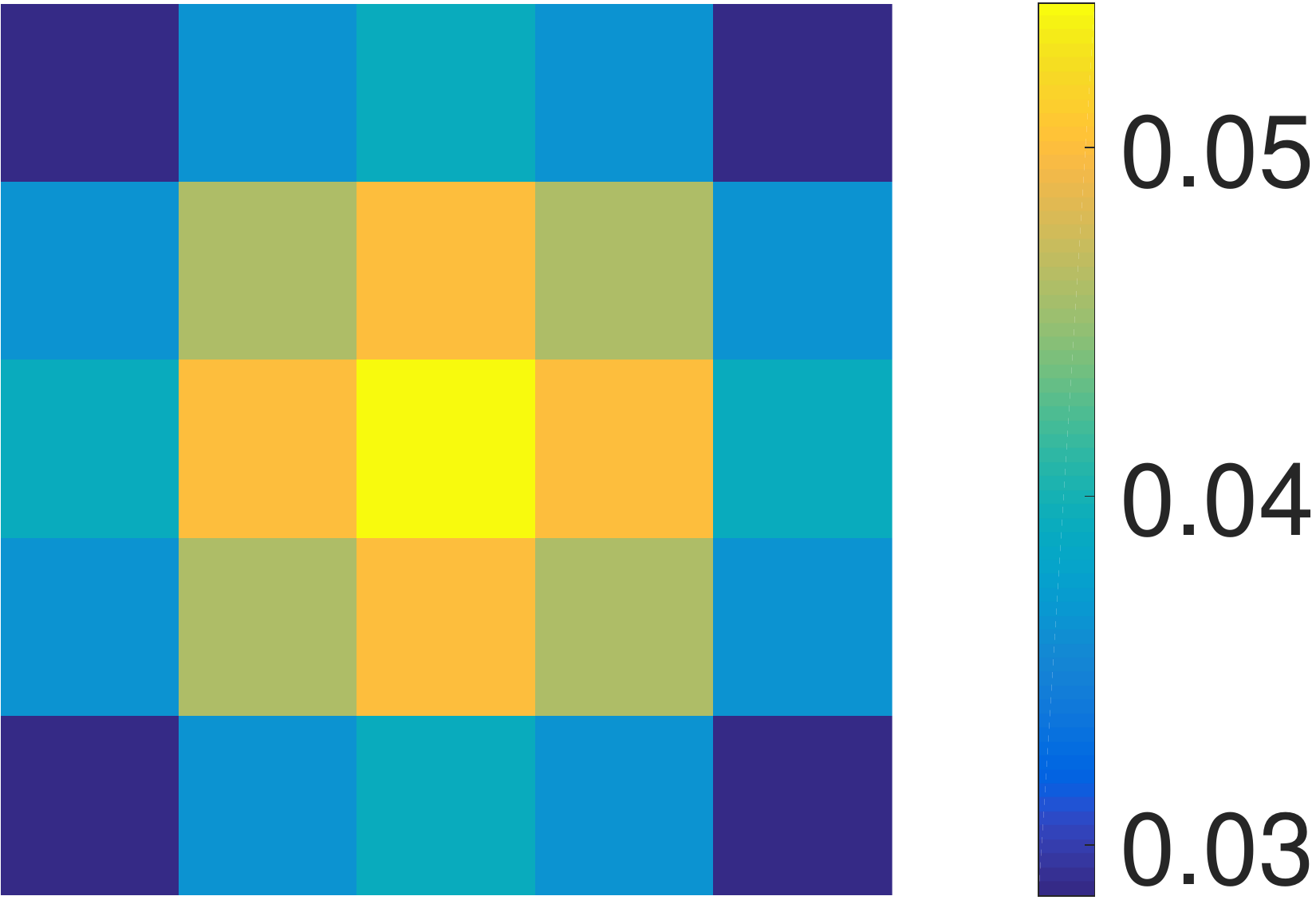}}
    \subfigure{
    \includegraphics[angle=0,width=0.24\textwidth]{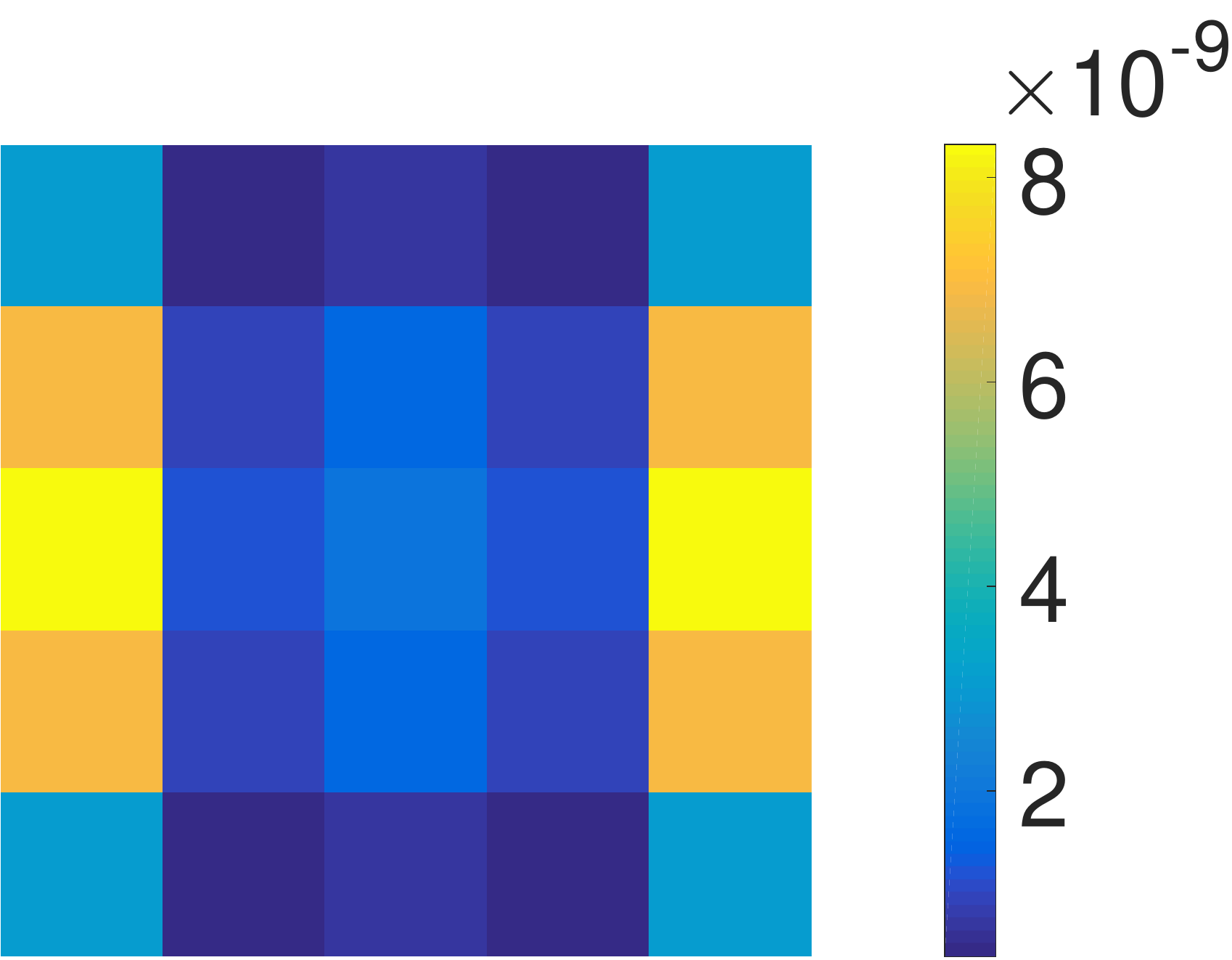}}
    \caption{Blurring kernel used in Section \ref{subsec:Fus_HS_MS} (left), in this section (middle) and their difference (right).}
\label{fig:kernel}
\end{figure}

In this section, we consider a kernel similar to the one used
in Section \ref{subsec:Fus_HS_MS}, which is displayed in the middle
of Fig. \ref{fig:kernel} (the difference between the two kernels is
shown in the right).
Note that this trivial change implies that the Fourier transform
of the new kernel has some values that are very close to zero,
which may drastically impact the performance of the FUSE algorithm.
The fusion performance of FUSE and R-FUSE with a TV prior is summarized in Table \ref{tb:robust_kernel}.
Obviously, the performance of FUSE degrades a lot due to the presence of close-to-zero values in the kernel FT,
which does not agree with the invertibility assumption of $\bfD$. On the contrary, the proposed R-FUSE  
provides results very close to (almost the same with) those obtained in Section \ref{subsec:Fus_HS_MS},
demonstrating its robustness w.r.t. the blurring kernel.

\begin{table}[t!]
\centering
\renewcommand{\arraystretch}{1.1}
\setlength{\tabcolsep}{0.5mm}
\caption{Performance of HS+MS fusion methods with a slightly different kernel: RSNR (in dB), UIQI, SAM (in degree), ERGAS, DD (in $10^{-3}$) and time (in second).}
\vspace{0.1cm}
\begin{tabular}{|c|c|cccccc|}
\hline
Prior & Methods & RSNR & UIQI & SAM  & ERGAS & DD & Time \\
\hline
\hline
\multirow{2}{*}{TV} &FUSE&  9.985  & 0.5640 & 14.50 & 8.348 &74.7  &133\\
\cline{2-8}
&R-FUSE &  29.629  & 0.9914 & 1.456 & 0.853 &6.761 &\Emp{115}\\
\hline
\end{tabular}
\label{tb:robust_kernel}
\end{table}
 \vspace{-0.2cm}
\section{Conclusion}
\label{sec:concls}
This paper developed a new robust and faster multi-band image fusion method
based on the resolution of a generalized Sylvester equation.
The application of the Woodbury formula allows any permutation 
in the frequency domain to be avoided and brings two benefits.
First, the invertibility assumption of the blurring operator is not necessary, 
leading to a more robust fusion strategy. 
Second, the computational complexity of the fusion algorithm
is reduced. Similar to the method in \cite{Wei2015FastFusion},
the proposed algorithm can be embedded into a block coordinate descent or 
an alternating direction method of multipliers
to implement (hierarchical) Bayesian fusion models. Numerical experiments confirmed 
that the proposed robust fast fusion method has the advantage of reducing the
computational cost and also is more robust to the blurring kernel conditioning,
compared with the  method investigated in \cite{Wei2015FastFusion}.

\bibliographystyle{ieeetran}
\bibliography{strings_all_ref,biblio_all}
\end{document}